\documentclass[10pt,twocolumn,letterpaper]{article}

\usepackage{cvpr}
\usepackage{times}
\usepackage{epsfig}
\usepackage{graphicx}
\usepackage{amsmath}
\usepackage{amssymb}
\usepackage{multirow}
\usepackage{adjustbox} 

\DeclareMathOperator*{\argmin}{arg\,min}
\usepackage[caption=false]{subfig} 
\usepackage[pagebackref=true,breaklinks=true,letterpaper=true,colorlinks,bookmarks=false]{hyperref}

\usepackage[activate={true,nocompatibility},final,tracking=true,kerning=true,spacing=true,factor=1100,stretch=10,shrink=9]{microtype}
\cvprfinalcopy 

\title{Neural Architecture Search for Deep Image Prior}
\def\cvprPaperID{3164} 

\newcommand{\squeezeupSmall}{\vspace{-2mm}}

\ifcvprfinal\fi
\begin{document}


\author{$^{[1]}$ Kary Ho, $^{[1]}$Andrew Gilbert, $^{[2]}$ Hailin Jin, $^{[1 2]}$ John Collomosse\\
$^{[1]}$ Centre for Vision Speech and Signal Processing, \\
University of Surrey \\
\and
$^{[2]}$ Creative Intelligence Lab, Adobe Research }


\maketitle

\begin{abstract}
  We present a neural architecture search (NAS) technique to enhance the performance of unsupervised image de-noising, in-painting and super-resolution under the recently proposed Deep Image Prior (DIP).  We show that evolutionary search can automatically optimize the encoder-decoder (E-D) structure and meta-parameters of the DIP network, which serves as a content-specific prior to regularize these single image restoration tasks.  Our binary representation encodes the design space for an asymmetric E-D network that typically converges to yield a content-specific DIP within 10-20 generations using a population size of $500$. The optimized architectures consistently improve upon the visual quality of classical DIP for a diverse range of photographic and artistic content.
  
\end{abstract}

\section{Introduction}

Many common image restoration tasks require the estimation of missing pixel data: de-noising and artefact removal; in-painting; and super-resolution.  Usually, this missing data is estimated from surrounding pixel data, under a smoothness prior.  Recently it was shown that the architecture of a randomly initialized convolutional neural network (CNN) could serve as an effective prior, regularizing estimates for missing pixel data to fall within the manifold of natural images.  This regularization technique, referred to as the {\em Deep Image Prior} (DIP) \cite{Ulyanov2018} exploits both texture self-similarity within an image and the translation equivariance of CNNs to produce competitive results for the image restoration tasks mentioned above.   However, the efficacy of DIP is dependent on the architecture of the CNN used; different content requires different CNN architectures for excellent performance and care over meta-parameter choices \eg filter sizes, channel depths, epoch count, etc.

\begin{figure}[t!]
    \centering
    \includegraphics[width=1.0\linewidth]{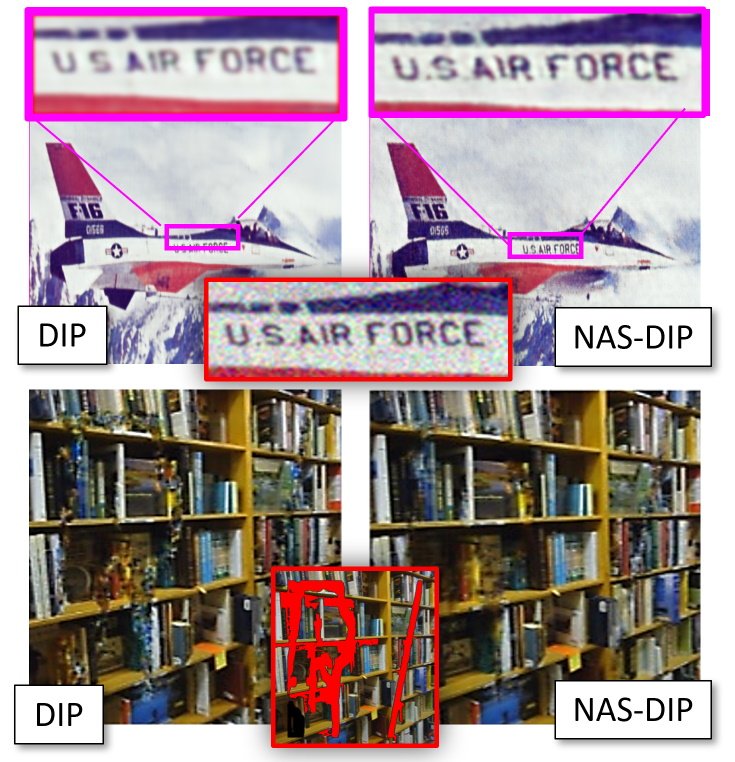}
    \caption{Neural Architecture Search yields a content-specific deep image prior (NAS-DIP, right) that enhances performance over classical DIP (left) for unsupervised image restoration tasks including de-noising (top) and  in-painting (bottom).  Source inset (red)
    } \label{fig:teaser}
    \squeezeupSmall
\end{figure}

This paper contributes an evolutionary strategy to automatically search for the optimal CNN architecture and associated meta-parameters given a single input image.  The core technical contribution is a genetic algorithm (GA) \cite{ea} for representing and optimizing a content-specific network architecture for use as DIP regularizer in unsupervised image restoration tasks. We show that superior results can be achieved through architecture search versus standard DIP backbones or random architectures.  We demonstrate these performance improvements for image de-noising, Content-Aware Fill (otherwise known as inpainting), and content up-scaling for static images; Fig.~\ref{fig:teaser} contrasts the output of classic DIP \cite{Ulyanov2018} with our neural architecture search (NAS-DIP).  

Unlike image classification, to which NAS has been extensively applied \cite{Real2017,Real2019,Liu2018,Elsken2019}, there is no ground truth available in the DIP formulation.  An encoder-decoder network (whose architecture we seek) is overfitted to reconstruct the input image from a random noise field, so acquiring a generative model of that image's structure. Parallel work in GANs \cite{DevilDecoder} has shown that architectural design, particularly of the decoder, is critical to learning a sufficient generative model, and moreover, that this architecture is content specific.  Our work is the first to both automate the design of encoder-decoder architectures and to do so without supervision, leveraging a state-of-the-art perceptual metric to guide the optimization \cite{LPIPS}.  Further, we demonstrate that the optimized architectures are content specific, and enable clustering on visual style without supervision. 

\section{Related Work}

 {\bf Neural architecture search (NAS)} seeks to automate the design of deep neural network architectures through data-driven optimization \cite{NASSurvey2019}, most commonly for image classification \cite{Zoph2018, Real2019}, but recently also object detection \cite{Zoph2018} and semantic segmentation \cite{Chen2018}.  NAS addresses a facet of the automated machine learning (AutoML) \cite{Hutter2019} problem, which more generally addresses hyper-parameter optimization and tuning of training meta-parameters.  

Early NAS leveraged Bayesian optimization for MLP networks \cite{Bergstra2013}, and was extended to CNNs \cite{Domhan2015} for CIFAR-10.  In their seminal work, Zoph and Le \cite{Zoph2017} applied reinforcement learning (RL) to construct image classification networks via an action space, tokenizing actions into an RNN-synthesised string, with reward driven by validation data accuracy.  The initially high computational overhead (800GPUs/4 weeks)  was reduced while further enhancing accuracy. For example, by exploring alternative  policies such as  proximal policy optimization \cite{Schulman2017} or Q-learning \cite{Baker2017}, RL approaches over RNN now scale to contemporary datasets \eg NASNet for ImageNet classification \cite{Negrinho2017,Zoph2018} and was recently explored for GAN over CIFAR-10 \cite{autogan}. Cai \etal similarly tokenizes the architecture, but explore the solution space via sequential transformation of the string via function-preserving mutation operations \cite{Cai2018} on an LSTM-learned embedding of the string.  Our work also encodes architecture as a (in our case, binary) string, but optimizes via an evolutionary strategy rather than training a sequential model under RL.  

Evolutionary strategies for network architecture optimization were first explored for MLPs in the early nineties \cite{Miller1989}.  Genetic algorithms (GAs) were used to optimize both the architecture and weights \cite{Angeline1994, Stanley2002}, rather than rely upon back-prop, in order to reduce the GA evaluation bottleneck; however, this is not practical for contemporary CNNs.  While selection and population culling strategies \cite{Elsken2019,Real2017,Real2019} have been explored to develop high performing image classification networks over ImageNet \cite{Liu2018, Real2019}.  Our work is unique in that we explore image restoration encoder-decoder network architectures via GA optimization, and as such, our architecture representation differs from prior work.

 {\bf Single image restoration} has been extensively studied in classical vision and deep learning, where priors are prescribed or learned from representative data.  A common prior to texture synthesis is the Markov Random Field (MRF) in which the pair-wise term encodes spatial coherence. Several in-painting works exploit MRF formulations of this kind~\cite{Kwatra2003,He2012,Liu2013}, including methods that source patches from the input \cite{Efros1999} or multiple external \cite{hays2007scene,Gilbert2018} images, or use random propagation of a few good matched patches~\cite{Barnes2009}. Patch self-similarity within single images has also been exploited for single image super-resolution \cite{GlasnerICCV} and de-noising.   The Deep Image Prior (DIP) \cite{Ulyanov2018} (and its video extension \cite{Zhang2019}) exploit translation equivariance of CNNs to learn and transfer patches within the receptive field. Very recently, single image GAN \cite{singan} has been proposed to learn a multi-resolution model of appearance and DIP has been applied to image layer decomposition \cite{Gandelsman2018}.   
 Our work targets the use cases for DIP proposed within the original paper \cite{Ulyanov2018}; namely super-resolution, de-noising and region in-painting.  All three of these use cases have been investigated substantially in the computer vision and graphics literature, and all have demonstrated significant performance benefits when solved using a deep neural network trained to perform that task on representative data. Generative Adversarial Networks (GANs) are more widely used for in-painting and super-resolution \cite{stackgan, Wang2018} by learning structure and appearance from a representative corpus image data~\cite{yeh2016semantic}, in some cases explicitly maintaining both local and global consistency through independent models ~\cite{IizukaSIGGRAPH2017}. Our approach and that of DIP differs in that we do not train a network to perform a specific task. Instead, we use an untrained (randomly initialized) network to perform the tasks by overfitting such a network to a single image under a task-specific loss using neural architecture search.    

\section{Architecture Search for DIP}

The core principle of DIP is to learn a generative CNN $G_\theta$ (where $\theta$ are the learned network parameters \eg weights) to reconstruct $x$ from a noise field $\mathcal{N}$ of identical height and width to $x$, with pixels drawn from a {\em uniform} random distribution.  Ulyanov \etal \cite{Ulyanov2018} propose a symmetric encoder-decoder network architecture with skip connections for $G_\theta$, comprising five pairs of (up-)convolutional layers with varying architectures depending on the image restoration application (de-noising, in-painting or super-resolution) and the image content to be processed.  A reconstruction loss is applied to learn $G_\theta$ for a single given image $x$:
\begin{equation}
\theta^* = \argmin_{\theta} \| G_{\theta}(\mathcal{N})-x\|^2_2
\label{eq:dip}
\end{equation}

Our core contribution is to optimize not only for $\theta$ but also for  architecture $G$ using a genetic algorithm (GA), guided via a perceptual quality metric \cite{LPIPS}, as now described.

\subsection{Network Representation}
\label{sec:archrep}
\begin{figure*}[t!]
\begin{center}
   \includegraphics[width=1\linewidth,height=6.5cm]{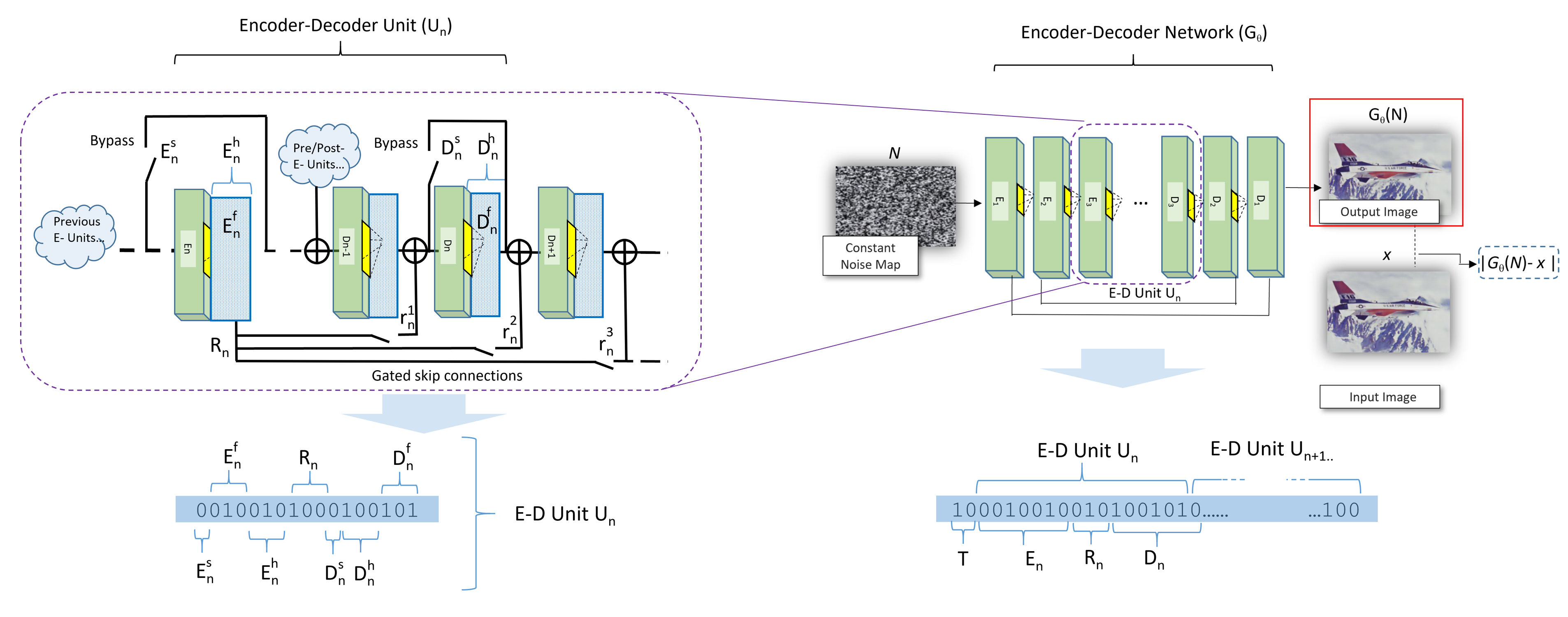}
\end{center}
   \caption{Architecture search space of NAS-DIP(-T).  The Encoder-Decoder (E-D) network $G$ (right) is formed of several E-D Units ($U_n$) each with an Encoder $E_n$) and Decoder $D_n$ paired stage (zoomed, left) represented each by 7 bits plus an additional $4N$ bits $R_n$ to encode gated skip connections from $E_n$ to other decoder blocks in the network. Optionally the training epoch count $T$ is encoded.  The binary  representation for E-D units is discussed further in Sec.~\ref{sec:archrep}.  Under DIP images are reconstructed from constant noise field $N$ by optimizing to find weights $\theta$ thus overfitting the network to input image $x$ under reconstruction loss \eg here for de-noising (eq.\ref{eq:tl1}).}
\label{fig:arch}
\end{figure*}

We encode the space of encoder-decoder (E-D) architectures from which to sample $G$ as a constant length binary sequence, representing $N$ paired encoder decoder units $U=\{U_1,...,U_N\}$. 
Following \cite{Ulyanov2018}, $G$ is a fully convolutional E-D network and optimize for the connectivity and meta-parameters of the convolutional layers. A given unit $U_n$ comprises encoder $E_n$ and decoder $D_n$ convolutional stages denoted $E_n$ and $D_n$ respectively each of which requires 7 bits to encode its parameter tuple.  Unit $U_n$ requires a total $14+4N$ bits to encode, as an additional $4N$-bit block for the unit, encodes a tuple specifying the configuration of skip connections from its encoder stage to each of the decoder stages \ie both within itself and other units.  Thus, the total binary representation for an architecture in neural architecture search for DIP ({\bf NAS-DIP}) is $N(14+4N)$. For our experiments, we use $N=6$ but note that the effective number of encoder or decoder stages varies according to skip connections which may bypass the stage.  Fig.~\ref{fig:arch} illustrates the organisation of the E-D units (right) and the detailed architecture of a single E-D unit (left).  Each unit $U_n$ comprises the following binary representation, where super-scripts indicate elements of the parameter tuple:

\begin{itemize}
    \item{$\mathrm{E}_n^s \in [0,1]$ (1 bit) a binary indicator of whether the encoder stage of unit $U_n$ is skipped (bypassed).}
    \item{$\mathrm{E}_n^f \in [0,7]$ (3 bits) encoding filter size $f=2E_n^f+1$ learned by the convolutional encoder.}
    \item{$\mathrm{E}_n^h \in [0,7]$ (3 bits) encoding number of filters $h=2^{E_n^h-1}$ and so channels output by the encoder stage.}
    \item{$\mathrm{D}_n^s \in [0,1]$ (1 bit) a binary indicator of whether the decoder stage of unit $U_n$ is skipped (bypassed).}
    \item{$\mathrm{D}_n^f \in [0,7]$ (3 bits) encoding filter size $f=2D_n^f+1$ learned by the up-convolutional decoder stage.}
    \item{$\mathrm{D}_n^h \in [0,7]$ (3 bits) encoding number of filters $h=2^{D_n^h-1}$ and so channels output by the decoder stage.}
    \item{$\mathrm{R}_n \in \mathbb{B}^{4N}$ (4N bits) encodes gated skip connections as $[\rho_n^1,...,\rho_n^N]$; each 4-bit group $\rho_n^i \in [0,15]$ determines whether gated skip path $r_n^i$ connects from $E_n$ to $D_i$ and if so, how many filters/channels (\ie skip gate is open if $r_n^i=0$).}
\end{itemize}

{\bf NAS-DIP-T.} We explore a variant of the representation encoding maximum epoch count $T=500*(2^t-1)$ via two additional bits coding for $t$ and thus a representation length for NAS-DIP-T of $N(14+4N)+2$.  

{\bf Symmetric NAS-DIP.}  We also explore a compact variant of the representation that forces $E_n=D_n$ for all parameters, forcing a symmetric E-D architecture to be learned and requiring only 10 bits to encode $U_n$.  However, we later show that such symmetric architectures are almost always underperformed by asymmetric architectures discovered via our search (subsec.~\ref{sec:evalvariants}).

\subsection{Evolutionary Optimization}
\begin{figure*}[t!]
    \centering
    \includegraphics[width=1.0\linewidth]{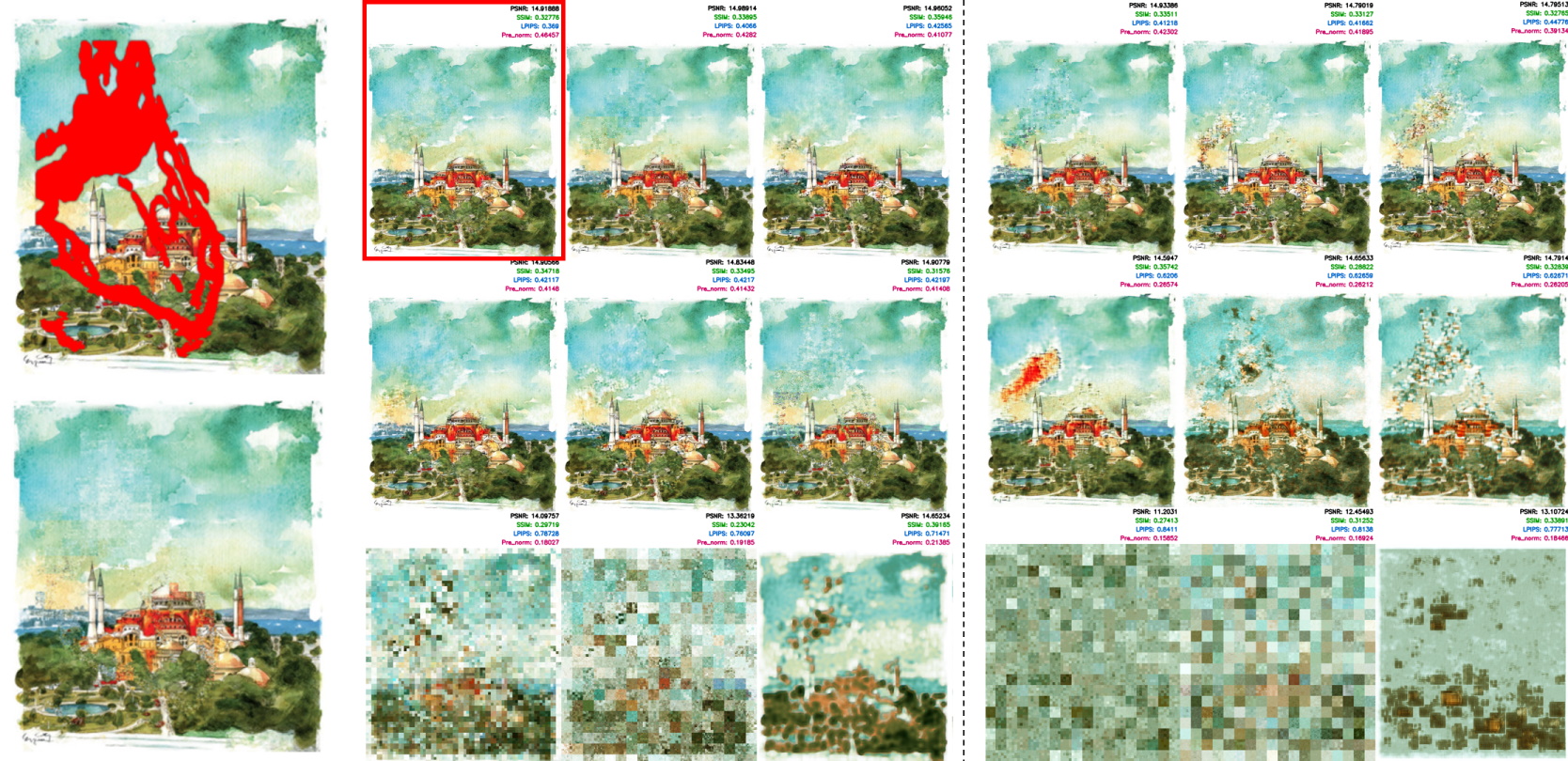}
    \caption{NAS-DIP convergence for in-painting task on BAM! \cite{wilber2017bamICCV}. Left: Input (top) and converged (bottom) result at generation 13.   Right: Sampling the top, middle and bottom performing architectures (shown in respective rows) from a population of 500 architectures, at the final (1st, middle) and final (13th, rightmost) generations.  Inset scores: Fitness selection is driven using LPIPS \cite{LPIPS}; evaluation by PSNR/SSIM.
    } \label{fig:libgen}
\end{figure*}

DIP provides an architecture specific prior that regularises the reconstruction of a given source image $x$ from a uniform random noise field $\mathcal{N}$.  Fixing $N$ constant, we use a genetic algorithm (GA) to search architecture space for the optimal architecture $G^*$ to recover a 'restored' \eg denoised, inpainted or upscaled version $\hat{x}$ of that  source image:
\begin{equation}
\hat{x} = G^*_{\theta^*}(\mathcal{N}).
\end{equation}
 GAs simulate the process of natural selection by breeding successive generations of individuals through the processes of cross-over, fitness-proportionate reproduction and mutation. In our implementation, such individuals are network configurations encoded via our binary representation.

Individuals are evaluated by running a pre-specified DIP image restoration task using the encoded architecture. We consider unsupervised or `blind'  restoration tasks (de-noising, in-painting, super-resolution) in which an ideal $\hat{x}$ is unknown (is sought) and so cannot guide the GA search.  In lieu of this ground truth we employ a perceptual measure (subsec.~\ref{sec:fitness}) to assess the visual quality generated by any candidate architecture by training $G_\theta^*(N)$ via backpropogation to minimize a task specific reconstruction loss:
\begin{eqnarray}
\mathcal{L}_{de-noise}(x;G) &=& min_\theta |G_\theta(\mathcal{N})-x|. \label{eq:tl1}\\
\mathcal{L}_{in-paint}(x;G) &=& min_\theta |M(G_\theta(\mathcal{N}))-M(x)|. \label{eq:tl2}\\
\mathcal{L}_{upscale}(x;G) &=& min_\theta |D(G_\theta(\mathcal{N}))-x|. \label{eq:tl3}
\end{eqnarray}

where $D$ is a downsampling operator reducing its target to the size of $x$ via bi-linear interpolation, and $M(.)$ is a masking operator that returns zero within the region to be in-painted. 

We now describe a single iteration of the GA search, which is repeated until the improvements gained over the previous few generations are marginal (the change in both average and maximum population fitness over a sliding time window fall below a threshold).

\subsubsection{Population Sampling and Fitness}
\label{sec:fitness}
We initialize a population of $K=500$ solutions uniformly sampling $\mathbb{B}^{N(14+4N)}$ to seed initial architectures $\mathcal{G}=\{G_1,G_2,...,G_K\}$.  The visual quality of $\hat{x}$ under each architecture is assessed `blind’ using a learned perceptual measure (LPIPS score) \cite{LPIPS} as a proxy for individual fitness: 
\begin{equation}
f(G_i) = \mathrm{P}( argmin_{\hat{x}} \mathcal{L}(\hat{x};G_\theta))
\label{eq:fit}
\end{equation}
where $\mathcal{L}$ is the task specific loss (Eq.\ref{eq:tl1}-\ref{eq:tl3}) and $P(.)$ is the perceptual score \cite{LPIPS} for a given individual $G_i$ in the population. We explored several blind perceptual scores from the GAN literature including Inception Score \cite{Salimans2016} and Frechet Inception Distance \cite{FID} but found LPIPS to better correlate with PSNR and SSIM during our evaluation, improving convergence for NAS-DIP.

Individuals are selected stochastically (and with replacement) for breeding, to populate the next generation of solutions.  Architectures that produce higher quality outputs are more likely to be selected. However, population diversity is encouraged via the stochasticity of the process and the introduction of a random mutation into the offspring genome. We apply elitism; the bottom 5\% of the population is culled, and the top 5\% passes unperturbed to the next generation -- the fittest individual in successive generations is thus at least as fit as those in the past. The middle 90\% is used to produce the remainder of the next generation.  Two individuals are selected stochastically with a bias to fitness $p(G_i) =  f(G_i)/{\sum_{j=1}^{K} f(G_j)}$, and bred via cross-over and mutation (subsec.~\ref{sec:crossover}) to produce a novel offspring for the successive generation. This process repeats with replacement until the population count matches the prior generation.

\begin{figure*}[t!]
\begin{center}
   \includegraphics[width=0.95\linewidth]{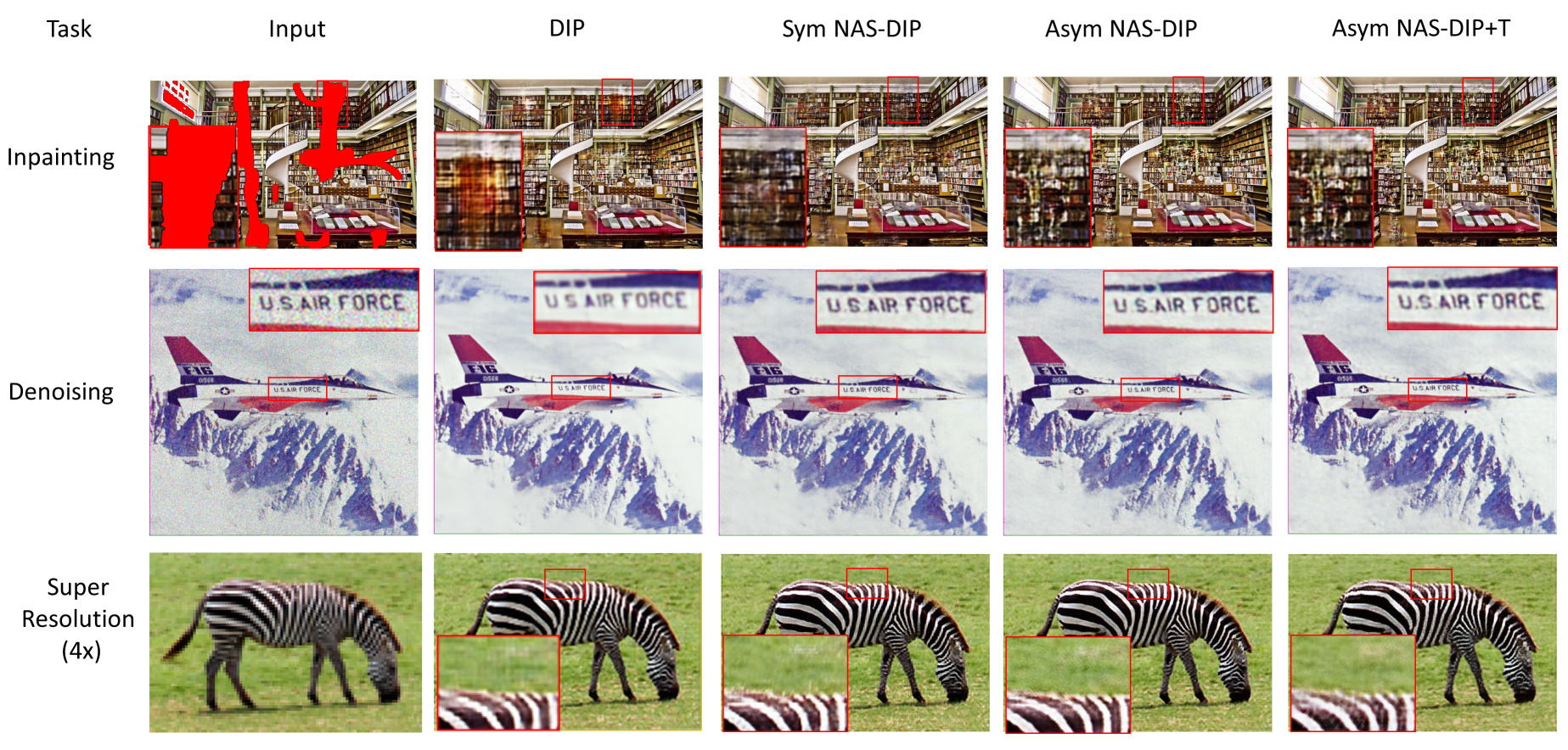}
   \includegraphics[width=0.9\linewidth]{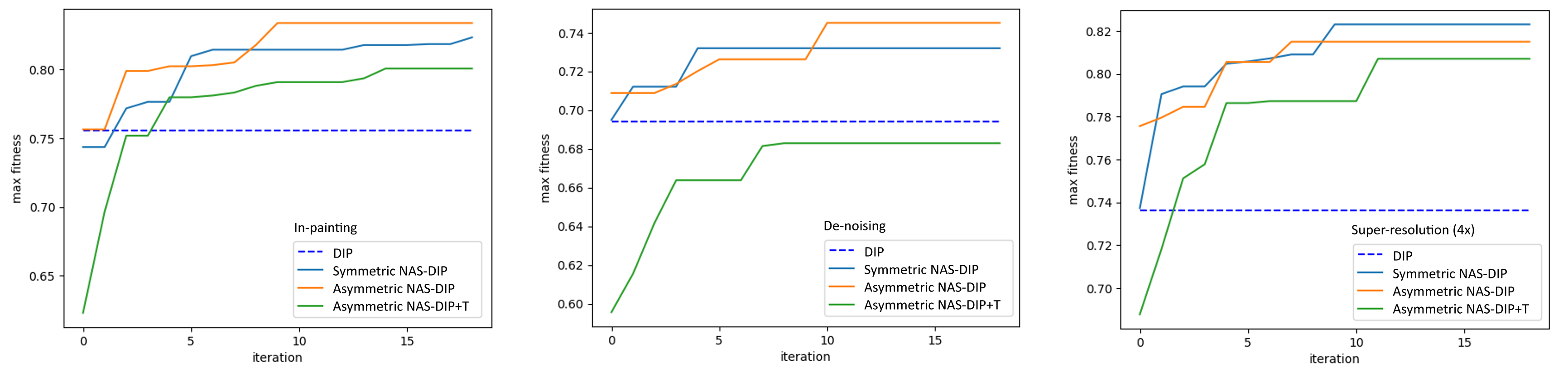}
\end{center}
   \caption{Evaluation of proposed NAS-DIP (and variants) vs. classical DIP \cite{Ulyanov2018} for in-painting, de-noising and $4 \times$ super-resolution on the Library, Plane and Zebra images of the {\em DIP} dataset.  For each task, respectively, from left to right, the max population fitness graph is shown.  We compare the proposed unconstrained (asymmetric) E-D network search of NAS-DIP (optionally with epoch count $T$ freely optimized; NAS-DIP-T) with NAS-DIP constrained to symmetric E-D architectures only.  The baseline (dotted blue) for classical DIP uses the architecture published in \cite{Ulyanov2018}.  Examples further  quantified in Table~\ref{tab:PerceptualMeasurementScores}.
   }
\label{fig:IndividualApproachPerformance}
\end{figure*}

\subsubsection{Cross-over and Mutation}
\label{sec:crossover}

Individuals are bred via genetic crossover; given two constant length binary genomes of form $G\{(E_1,R_1,D_1),...,(E_N,R_N,D_N)\} \in \mathbb{B}^{N(14+4N)}$, a splice point $S=[1,N]$ is randomly selected such that units from a pair of parents $A$ and $B$ are combined via copying of units $U_{N<S}$ from $A$ and $U_{N \geq S}$ from B. Such cross-over could generate syntactically invalid genomes \eg due to tensor size incompatibilities between units in the genome. During the evaluation, an invalid architecture evaluates to zero prohibiting its selection for subsequent generations.

Population diversity is encouraged via the stochasticity of the selection process and the introduction of a random mutation into the offspring genome. Each bit within the offspring genome is subject to random flip with low probability $p_m$; we explore the trade-off of this rate against convergence in subsec.~\ref{sec:evalmut}.

\section{Experiments and Discussion}
\label{sec:eval}

We evaluate the proposed neural architecture search technique for DIP (NAS-DIP) for each of the three blind image restoration tasks proposed originally for DIP \cite{Ulyanov2018}: image in-painting, super-resolution and de-noising.  

{\bf Datasets.} We evaluate over three public datasets:  1) Places2~\cite{zhou2016places}; a dataset of photos commonly used for in-painting (test partition sampled as~\cite{IizukaSIGGRAPH2017}); 2) Behance Artistic Media (BAM!) \cite{bam}; a dataset of artwork, using the test partition of BAM! selected to evaluate in-painting in \cite{Gilbert2018} comprising of 8 media styles; 3) DIP \cite{Ulyanov2018}; the dataset of images used to evaluate the original DIP algorithm of Ulyanov \etal.  Where baseline comparison is made to images from the latter, the specific network architecture published in \cite{Ulyanov2018} is replicated using according to \cite{UlyanovJournal} and the authors' public implementation. Results are quantified via three objective metrics; PSNR (as in DIP) and structural similarity (SSIM)~\cite{wang2004imageSSIM} are used to evaluate against ground truth, and we also report the perceptual metric (LPIPS)~\cite{LPIPS} used as fitness score in the GA optimization.

\begin{table*}[t!]

\centering
\begin{tabular}{|l|c||c|c||c|c||c|c||}
\hline
\multirow{2}{*}{Dataset}   &\multirow{2}{*}{Task} &\multicolumn{2}{c||}{PSNR (higher better)}&\multicolumn{2}{c||}{LPIPS (lower better)}&\multicolumn{2}{c||}{SSIM (higher better)} \\ \cline{3-8} 
                             &          &DIP       &Proposed                       &DIP       &Proposed                       &DIP       &Proposed\\ \hline
BAM! \cite{wilber2017bamICCV}&In-painting& 16.8     &{\bf 19.76}                          &0.42      & {\bf 0.25}                          & 0.32     & {\bf 0.62}\\  
Places2 \cite{IizukaSIGGRAPH2017} &In-painting& 12.4     &{\bf 15.43}                          & 0.37     &{\bf 0.27}                           &  0.37    &{\bf 0.58} \\ \hline 
\multirow{3}{*}{DIP \cite{Ulyanov2018}}&In-painting& 23.7     &{\bf 27.3}                           &0.33      &{\bf 0.04}                           &.76       &{\bf 0.92}\\
                          &De-noising& 17.6     & {\bf 18.95}                         &0.13      &{\bf 0.09}                           &0.73      &{\bf 0.85}       \\
                          &Super Resolution&18.9&{\bf 19.3}                           &0.38      &{\bf 0.16}                           & 0.48     &{\bf 0.62}  \\ \hline
\end{tabular}

\caption{Per-dataset average performance of NAS-DIP (asymmetric representation) versus DIP over the dataset proposed in DIP, and for in-painting datasets BAM! and Places2.}
\label{tab:headlineResults}
\end{table*}

\begin{table*}[t!]
\begin{adjustbox}{width=1.0\textwidth}
\centering
\begin{tabular}{|l|l||c|c|c|c||c|c|c|c||c|c|c|c||}
\hline
\multirow{3}{*}{Task} &\multirow{3}{*}{Image}& \multicolumn{4}{c||}{PSNR (higher better)} & \multicolumn{4}{c||}{LPIPS (lower better)} & \multicolumn{4}{c||}{SSIM (higher better)} \\ \cline{3-14} 
                       &&\multirow{2}{*}{DIP}&Sym&Asym&Asym&\multirow{2}{*}{DIP}&Sym&Asym&Asym&\multirow{2}{*}{DIP}& Sym&Asym&Asym \\ 
                           &        &     & NAS-DIP&NAS-DIP&NAS-DIP-T      &     &NAS-DIP &NAS-DIP       &NAS-DIP-T&     &NAS-DIP     &NAS-DIP       &NAS-DIP-T\\ \hline
\multirow{3}{*}{In-painting}& Vase   &18.3&29.8&29.2&\textbf{30.2}          &0.76&0.02    &\textbf{0.01}&0.02   &0.48  &0.95       &\textbf{0.96} &0.95\\
                           & Library&19.4& 19.7&\textbf{20.4}&20.0         &0.15 & 0.10  &\textbf{0.09}&0.12     &0.83 & 0.81       &\textbf{0.84} &0.83\\
                           & Face   &33.4& 34.2&\textbf{36.0}&35.9&0.078 &0.03 &\textbf{0.01}&0.04    &0.95 & 0.95       &\textbf{0.96} &0.95\\ \hline
\multirow{2}{*}{De-noising} & Plane  &23.1&25.8& 25.7&\textbf{25.9}&0.15 & 0.096  &0.09&\textbf{0.07}    &0.85 & 0.90       &0.92 &\textbf{0.94}\\  
                           & Snail  &12.0&12.6& 12.2        &\textbf{12.7} &0.11 & 0.12   &0.09&\textbf{0.06}    &0.61 & 0.52       &0.74 &\textbf{0.84}\\ \hline
\multirow{2}{*}{Super-Resolution}& Zebra 4x & 19.1&\textbf{20.2}&19.6&20.0 &0.19 & \textbf{0.13}   &0.14&\textbf{0.14}     &0.67 & 0.51       &\textbf{0.75} &0.72\\  
                           & Zebra 8x& 18.7 &\textbf{19.1}&18.96     &\textbf{19.1} &0.57 & 0.29&\textbf{0.19}&0.20&0.28& 0.34      &\textbf{0.48} &0.47  \\ \hline
\end{tabular}

\end{adjustbox}
\caption{Detailed quantitative comparison of visual quality under three image restoration tasks, for the DIP dataset.  Comparison between the architecture presented in the original DIP for each image, with the best architecture found under each of three variants of the proposed method: NAS-DIP, NAS-DIP-T and NAS-DIP constrained to symmetric E-D (subsec.~\ref{sec:archrep}).  Quality metrics: LPIPS \cite{LPIPS} (used in NAS objective); and both PSNR and SSIM~\cite{wang2004imageSSIM} as external metrics.}
\label{tab:PerceptualMeasurementScores}
\end{table*}

\begin{figure*}[t!]
\small
a) \includegraphics[width=0.55\linewidth,height=5cm]{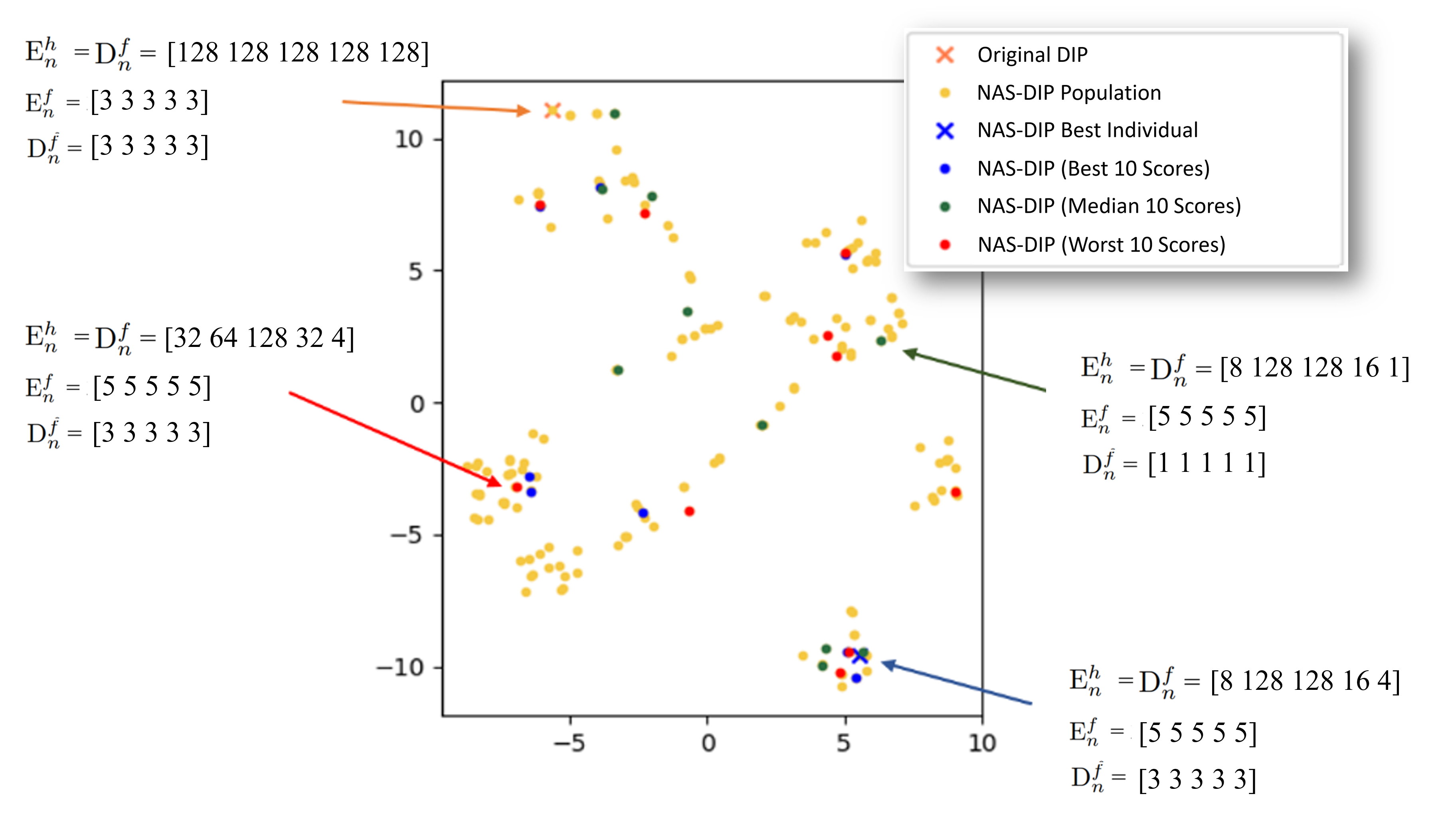}~~~~~
b) \includegraphics[width=0.4\linewidth,height=5cm]{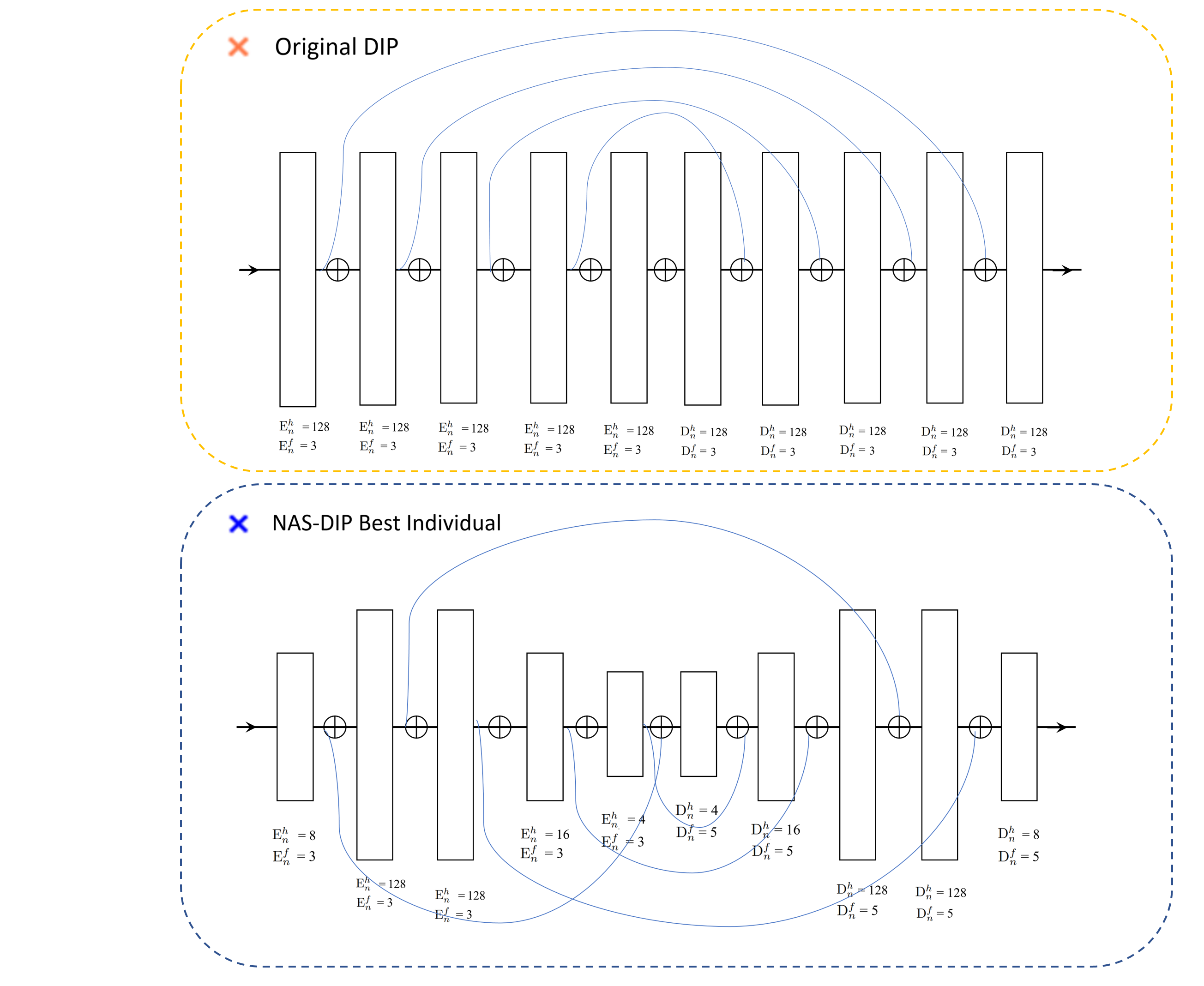}
   \caption{t-SNE visualizations in architecture space ($\mathbb{B}^{N(14+4N)}$) for NAS-DIP, $N=5$ symmetric network: a) Population distribution at convergence; multi-modal distribution of similar best DIP architectures obtained for the super-resolution task (image inset); b) Visualizing best discovered architecture from the population (lower), and the classical DIP default architecture \cite{UlyanovJournal} (upper) for this task.  }
\label{fig:tsne}
\end{figure*}

{\bf Training details.} We implement NAS-DIP in Tensorflow, using ADAM and learning rate of $10^{-3}$ for all experiments.  For NAS-DIP epoch count is 2000 otherwise this and other metaparameters are searched via the optimization. To alleviate the evaluation bottleneck in NAS (which takes, on average, 2-3 minutes per architecture proposal), we distribute the evaluation step over 16 Nvidia Titan-X GPUs capable of each evaluating two proposals concurrently for an average NAS-DIP search time of 3-6 hours total (for an average of 10-20 generations to convergence).  Our DIP implementation extends the authors' public code for DIP \cite{Ulyanov2018}.  Reproducibility is critical for NAS-DIP optimization; the same genome must yield the same perceptual score for a given source image. In addition to fixing all random seeds (\eg for batch sampling and fixing the initial noise field), we take additional technical steps to avoid non-determinism in cuDNN \eg avoidance of atomic add operations in image padding present in original DIP code.

\subsection{Network Representation}
\label{sec:evalvariants}

We evaluate the proposed NAS method over three variants of the architecture representation proposed in subsec.\ref{sec:archrep}: NAS-DIP, NAS-DIP-T (in which epoch count is also optimized), and a constrained version of NAS-DIP forcing a symmetric E-D network.  For all experiments, $N=5$ enabling E-D networks of up to 10 (up-)convolutional layers with gated skip connections. Performance is evaluated using PSNR for comparison with original DIP \cite{Ulyanov2018} as well as SSIM, and the LPIPS score used to guide NAS is also reported.  Table~\ref{tab:PerceptualMeasurementScores} provides direct comparison on images from Ulyanov \etal \cite{Ulyanov2018}; visual output and convergence graphs are shown in Fig.~\ref{fig:IndividualApproachPerformance}.  Following random initialization relative fitness gains of up to 30\% are observed after 10-20 generations after which performance converges to values above the DIP baseline \cite{Ulyanov2018} in all cases.

\begin{table*}[ht]
\begin{adjustbox}{width=1.0\textwidth}
\centering
\begin{tabular}{|l|l||c|c|c|c|c||c|c|c|c|c||c|c|c|c|c|}
\hline
 &  & \multicolumn{5}{c|}{PSNR (higher better)} & \multicolumn{5}{c|}{LPIPS (lower better)} & \multicolumn{5}{c|}{SSIM (higher better)} \\ \hline
Dataset                       &Image&DIP   &$p_m=0.01$&$p_m=0.02$&$p_m=0.05$&$p_m=0.10$&DIP  &$p_m=0.01$&$p_m=0.02$&$p_m=0.05$&$p_m=0.10$&DIP &$p_m=0.01$&$p_m=0.02$&$p_m=0.05$&$p_m=0.10$ \\  \hline
\multirow{3}{*}{In-painting}& Vase   &18.3&24.5         &28.5  &\textbf{30.2}&29.4      &0.76 &0.12      &0.10      &\textbf{0.01}&0.07   &0.48&0.57      &0.67      &\textbf{0.96}&0.69       \\
                           & Library&19.4  & 20.0     & 19.5     &\textbf{20.4}&19.8   &0.15 &0.12      &0.11      &\textbf{0.09}& 0.09  &0.83&0.81      &0.81      &\textbf{0.84}&0.82       \\
                           & Face   &33.4  & 34.5     & 35.6     &\textbf{35.9}&35.4   &0.01 &0.01      &0.01      &\textbf{0.01}&0.01   &0.95 &0.95     & 0.95     &\textbf{0.96}&0.95\\ \hline
\multirow{2}{*}{De-noising} & Plane  &23.1  &  22.9    &23.7      &\textbf{25.9}& 23.6  &0.15 &0.11      &0.01      &\textbf{0.09}&0.09   &0.85 & 0.85    &0.92      &\textbf{0.94}&0.91\\  
                           & Snail  &12.0  &12.0      &12.3      &\textbf{12.7}& 12.2  &0.11 &0.10      &0.09      &\textbf{0.00}&0.02   &0.61 & 0.78    &0.82      &\textbf{0.84}&0.81\\ \hline
\multirow{2}{*}{Super-Resolution}& Zebra 4x & 19.1&19.6&19.7     &\textbf{20.0}&19.6   &0.19 & 0.15     &0.15      &\textbf{0.14}&0.16   &0.67 & 0.70    & 0.70     &\textbf{0.72}&0.71\\  
                           & Zebra 8x& 18.7 & 18.9    & 19.0     &\textbf{19.1}&19.0   &0.57 & 0.34     &0.24      &\textbf{0.19}&0.28   &0.32 &0.35     &0.38      &\textbf{0.47}&0.43 \\ \hline
\end{tabular}

\end{adjustbox}
\caption{Quantitative results for varying mutation rates; chance of bit flip in offspring $p(r)$.  Comparing each of three variants of the proposed method: NAS-DIP, NAS-DIP-T and NAS-DIP constrained to symmetric E-D (subsec.~\ref{sec:archrep}) against original DIP.  Quality metrics: LPIPS \cite{LPIPS} (used in NAS objective); and both PSNR and SSIM as external metrics. }
\label{tab:mutation}
\end{table*}

\begin{figure*}[t]
\begin{center}
   \includegraphics[width=1\linewidth]{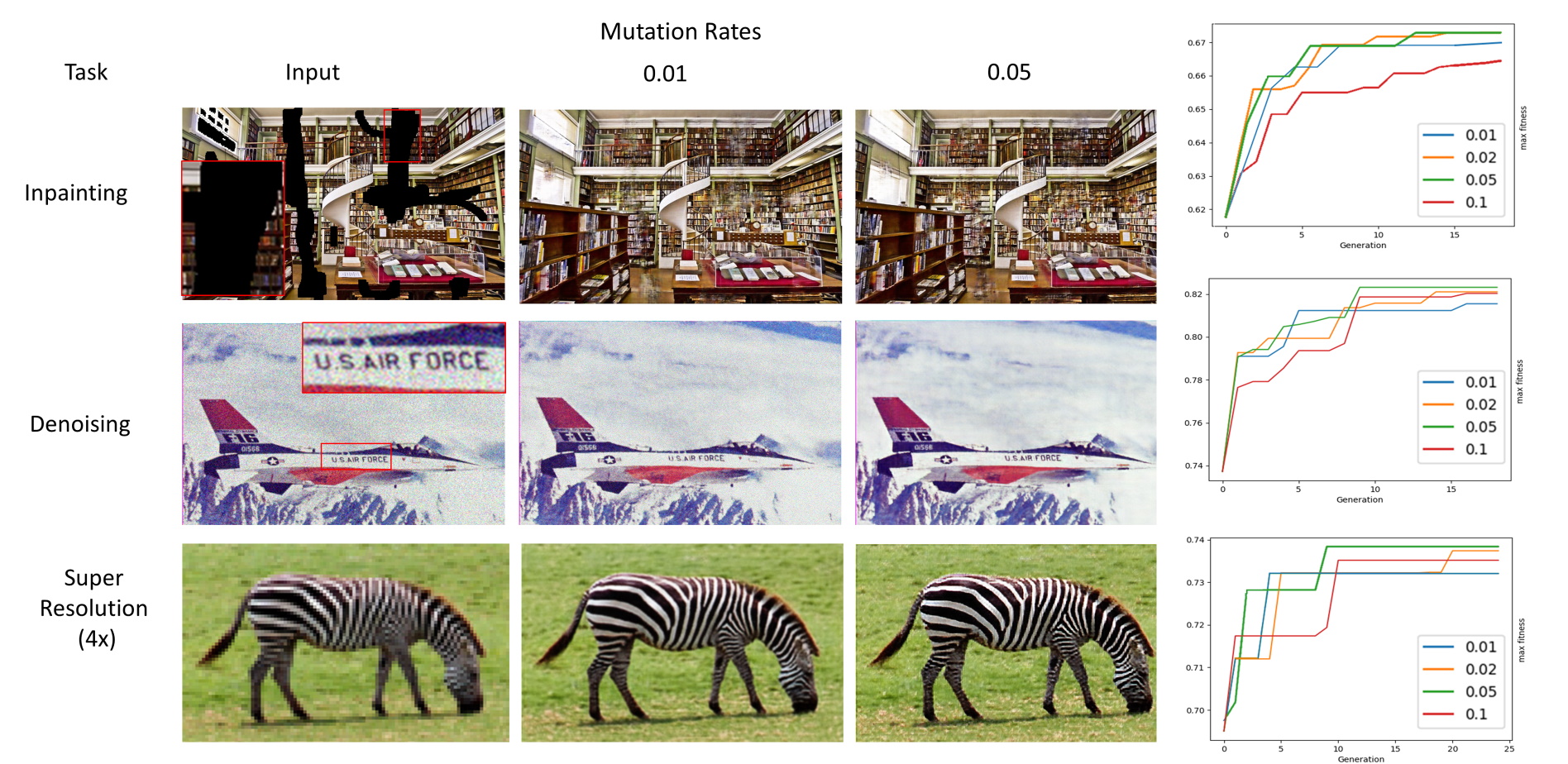}
\end{center}
   \caption{Result of varying mutation (bit flip) rate $p(r)$; improved visual quality (left, zoom recommended) is obtained with a mutation rate of $p(r)=0.05$ equating to an average of 4 bit flips per offspring. Convergence graph for each visual example (right).  }
\label{fig:MutationRateImgEx}
\end{figure*}

For both in-painting and de-noising tasks, asymmetric networks are found that outperform any symmetric network, including the networks published in the original DIP for those images and tasks.  For super-resolution, a symmetric network is found to outperform asymmetric networks and constraining the genome in this way aids in discovering a performant network. In all cases, it was unhelpful to optimize for epoch count $T$, despite prior observations on the importance of $T$ in obtained good reconstructions under DIP \cite{Ulyanov2018}.  Table~\ref{tab:headlineResults} broadens the experiment for NAS-DIP, averaging performance for three datasets: BAM! with 80 randomly sampled images, 10 from each 8 styles; Places2 with a 50 image subset included in~\cite{IizukaSIGGRAPH2017}; DIP with 7 images used in \cite{Ulyanov2018}.  For all three tasks and all three datasets, NAS-DIP discovers networks outperforming those of DIP \cite{Ulyanov2018}.

\begin{figure*}[t!]%
\captionsetup[subfigure]{labelformat=empty}
\centering
\subfloat[]{{\includegraphics[width=0.14\linewidth,height=2cm]{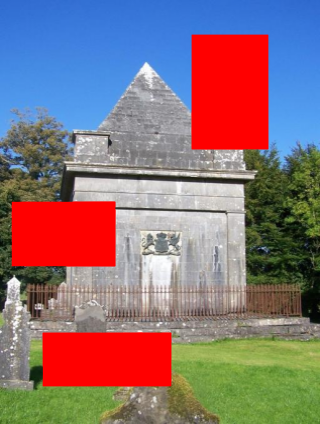} }}%
\subfloat[]{{\includegraphics[width=0.14\linewidth,height=2cm]{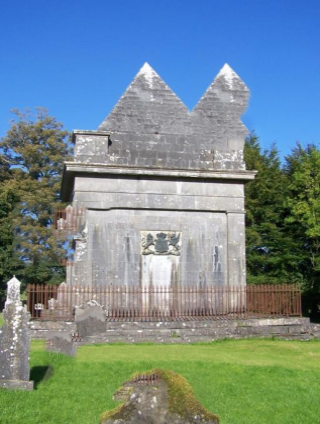} }}%
\subfloat[]{{\includegraphics[width=0.14\linewidth,height=2cm]{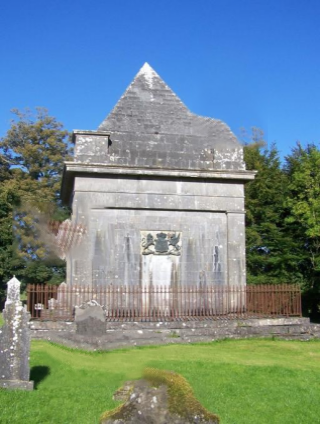} }}%
\subfloat[]{{\includegraphics[width=0.14\linewidth,height=2cm]{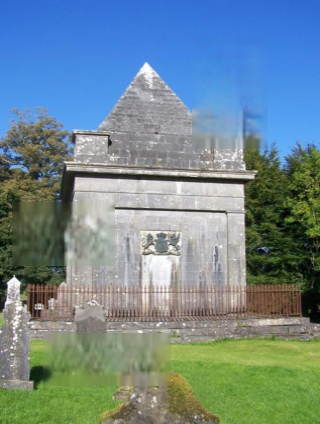} }}%
\subfloat[]{{\includegraphics[width=0.14\linewidth,height=2cm]{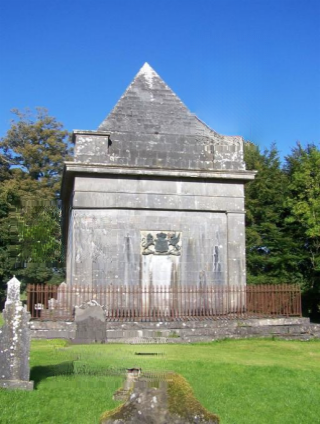} }}%
\subfloat[]{{\includegraphics[width=0.14\linewidth,height=2cm]{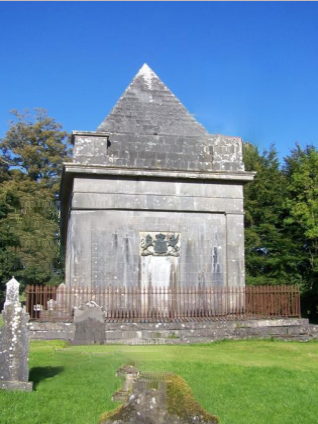} }} 
\subfloat[]{{\includegraphics[width=0.14\linewidth,height=2cm]{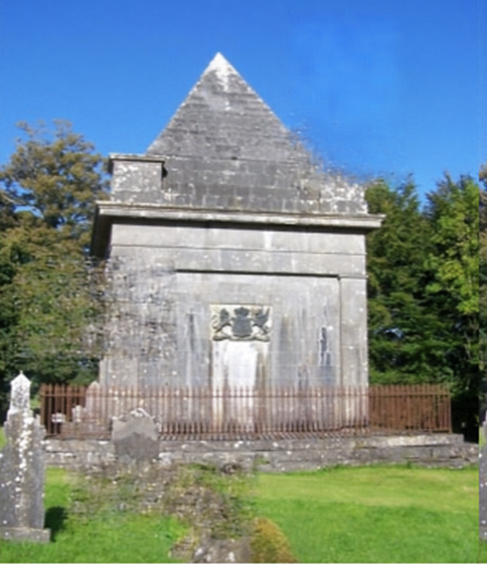} }} \\[-4ex]
\subfloat[Source / Mask (red)]{{\includegraphics[width=0.14\linewidth]{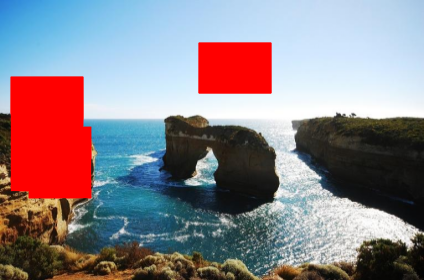} }}%
\subfloat[PatchMatch~\cite{Barnes2009}]{{\includegraphics[width=0.14\linewidth]{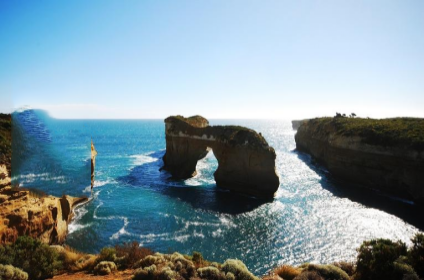} }}%
\subfloat[ImgMelding~\cite{darabi2012image}]{{\includegraphics[width=0.14\linewidth]{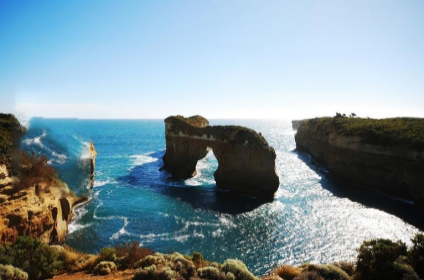} }}%
\subfloat[Context Encoder~\cite{pathakCVPR16context}]{{\includegraphics[width=0.14\linewidth]{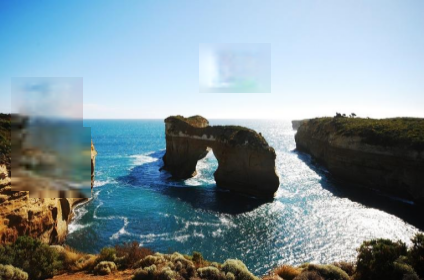} }}%
\subfloat[ImgComp~\cite{IizukaSIGGRAPH2017}]{{\includegraphics[width=0.14\linewidth]{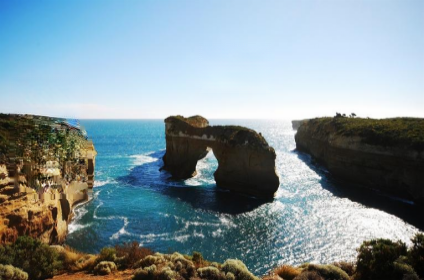} }}%
\subfloat[Style-aware~\cite{Gilbert2018}]{{\includegraphics[width=0.14\linewidth]{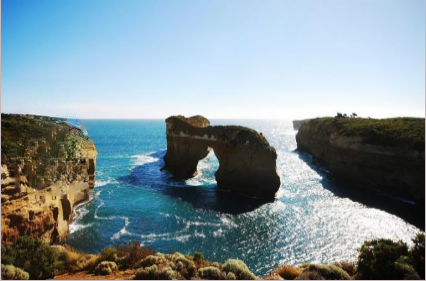} }}%
\subfloat[Proposed]{{\includegraphics[width=0.14\linewidth]{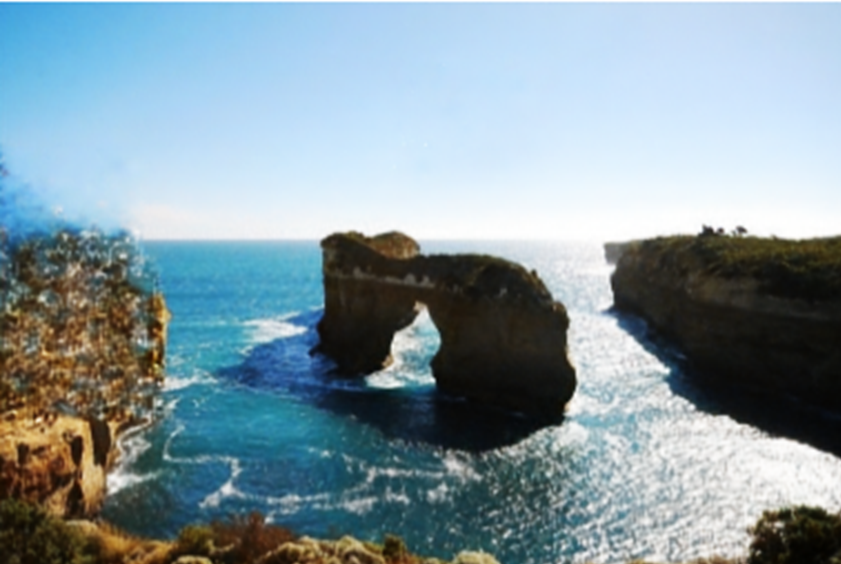} }}%

\caption{Qualitative visual comparison of NAS-DIP in-painting quality against 5 baseline methods; examples from dataset Places2 \cite{zhou2016places}.}%
\label{fig:Places}%
\end{figure*}
\subsection{Evaluating Mutation Rate}
\label{sec:evalmut}

\begin{figure}[t]
\small
   a)~~\includegraphics[width=0.9\linewidth]{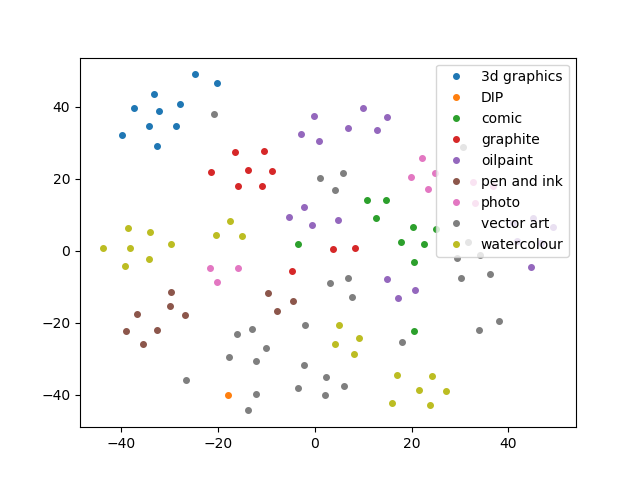}\\
   b)~~\includegraphics[width=0.9\linewidth,height=7cm]{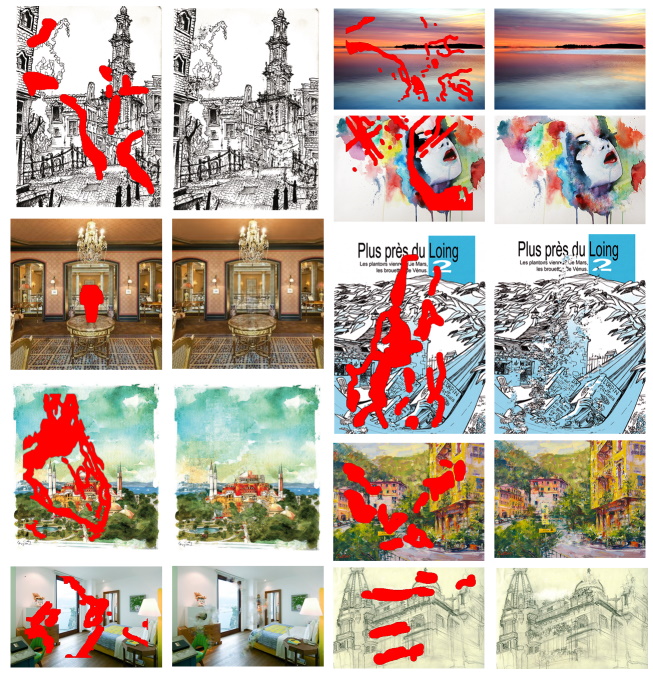}
   \caption{Content specific architecture discovery. NAS-DIP in-painted artwork results from BAM! \cite{wilber2017bamICCV} a) t-sne visualization of discovered architectures in $\mathbb{B}^{N(14+4N)}$ for 80 in-painted artworks (7 BAM styles + DIP photograph); common visual styles yield a common best performing architecture under NAS-DIP. b) Sample in-painted results (source/mask left, output right).}
\label{fig:initalResults}
\end{figure}

Population elitism requires a moderate mutation rate $p(r)$ to ensure population diversity and so convergence, yet raised too high convergence occurs at a lower quality level.  We evaluate  $p_m=\{0.01,0.02,0.05,0.10\}$ in Table~\ref{tab:mutation}, observing that for all tasks a bit flip probability of 5\% (\ie an average of 4 bit flips per offspring for $N=5$) encourages convergence the highest fitness value.  This in turn correlates to highest visual quality according to both PSNR and SSIM external metrics.  Fig.~\ref{fig:MutationRateImgEx} provides representation visual output alongside convergence graphs.  In general, convergence is achieved between 10-20 generations taking a few hours on single GPU for a constant population size $K=500$.


\subsection{Content aware in-painting}

To provide a qualitative performance of the Content-Aware in-painting we compare our proposed approach against several contemporary baselines: PatchMatch~\cite{Barnes2009} and Image Melding~\cite{darabi2012image} which sample patches from within a single source image; and two methods that use millions or training images, the generative convnet approach 'Context encoder'~\cite{pathakCVPR16context} a recent GAN in-painting work~\cite{IizukaSIGGRAPH2017}, and the Style-aware in-painting~\cite{Gilbert2018}. We compare against the same baselines using scenic photographs in Places2,  Fig~\ref{fig:Places} presents a visual comparison, and quantitative results are given in Tbl.\ref{tab:headlineResults} over the image set included in ~\cite{IizukaSIGGRAPH2017}  with the same mask regions published in that work.

\subsection{Discovered Architectures}

Architectures discovered by NAS-DIP are visualized in Fig.~\ref{fig:tsne}a, using t-SNE projection in the architecture space ($\mathrm{B}^{N(14+4N)}$) for a solution population at convergence (generation 20) for a representative super-resolution task for which symmetric E-D network was sought (source image inset).  Distinct clusters of stronger candidate architectures have emerged each utilising all 10 convolutional stages and similar filter sizes but with distinct skip activations and channel widths; colour coding indicates fitness.  Fig.~\ref{fig:tsne}b shows a schematic for the default architecture for this task published in \cite{UlyanovJournal} alongside the discovered best architecture for this task and image.

Fig.~\ref{fig:initalResults} visualizes 80 best performing networks for in-painting BAM! artworks of identical content (flowers) but exhibiting 8 different artistic styles.  Each image has been run through NAS-DIP under loss eq.~\ref{eq:tl2}; representative output is included in Fig.~\ref{fig:initalResults}.  Visualizing the best performing architectures via t-SNE in $\mathrm{B}^{N(14+4N)}$ reveals style-specific clusters; pen-and-ink, graphite sketches, 3D graphics renderings, comics, and vector artwork all form distinct groups while others \eg watercolor form several clusters.  We conclude that images that share a common visual style exhibit commonality in network architecture necessary perform image reconstruction well.  This discovery suggests a potential application for NAS-DIP beyond architecture search to unsupervised clustering \eg for visual aesthetics.

\section{Conclusion}

We reported the first neural architecture search (NAS) for image reconstruction under the recently proposed deep image prior (DIP) \cite{Ulyanov2018}, learning a content-specific prior for a given source image in the form of an encoder-decoder (E-D) network architecture.  Following the success of evolutionary search techniques for image classification networks \cite{Real2017,Real2019} we leveraged a genetic algorithm to search a binary space representing asymmetric E-D architectures and demonstrated its efficacy for image de-noising, in-painting and super-resolution.  For the latter case, we observed a constrained version of our genome yielding symmetric networks exceeded that of asymmetric networks which benefited the other two tasks. All image restoration tasks were `blind' and optimization was guided via a proxy measure for perceptual quality \cite{LPIPS}. In all cases, we observed the performance of the discovered networks to significantly exceed classical DIP and the potential for content-specific architectures beyond image restoration to unsupervised style clustering.    Future work could pursue the latter parallel application, as well as explore further generalizations of the architecture space beyond fully convolutional E-Ds to incorporate pooling and normalization strategies. 

{\small
\bibliographystyle{ieee_fullname}


\begin{thebibliography}{10}\itemsep=-1pt

\bibitem{Angeline1994}
P. Angeline, G. Saunders, and J. Pollack.
\newblock An evolutionary algorithm that constructs recurrent neural networks.
\newblock {\em IEEE Trans. on Neural Networks}, 5(1):54--65, 1994.

\bibitem{Baker2017}
B. Baker, O. Gupta, N. Naik, and R. Raskar.
\newblock Designing neural network architectures using reinforcement learning.
\newblock In {\em Proc. ICLR}, 2017.

\bibitem{Barnes2009}
C. Barnes, E. Shechtman, A. Finkelstein, and D. Goldman.
\newblock Patchmatch: a randomized correspondence algorithm for structural
  image editing,.
\newblock In {\em Proc. ACM SIGGRAPH}, 2009.

\bibitem{Bergstra2013}
J. Bergstra, D. Yamins, and D. Cox.
\newblock Making a science of model search: Hyper-parameter optimization in
  hundreds of dimensions for vision architectures.
\newblock In {\em Proc. ICML}, 2013.

\bibitem{Cai2018}
H. Cai, T. Chen, W. Zhang, Y. Yu, and J. Wang.
\newblock Efficient architecture search by network transformation.
\newblock In {\em Proc. AAAI}, 2018.

\bibitem{Chen2018}
L. Chen, M. Collins, Y. Zhu, G. Papandreou, B. Zoph, F. Schroff, H. Adam, and
  J. Shlens.
\newblock Searching for efficient multi-scale architectures for dense image
  prediction.
\newblock In {\em Proc. NeurIPS}, pages 8713--8724, 2018.

\bibitem{darabi2012image}
Soheil Darabi, Eli Shechtman, Connelly Barnes, Dan~B Goldman, and Pradeep Sen.
\newblock Image melding: combining inconsistent images using patch-based
  synthesis.
\newblock In {\em ACM TOG}, 2012.

\bibitem{ea}
K. de Jong.
\newblock Learning with genetic algorithms.
\newblock {\em J. Machine Learning}, 3:121--138, 1988.

\bibitem{Domhan2015}
T. Domhan, J.~T. Springenberg, and F. Hutter.
\newblock Speeding up automatic hyperparameter optimization of deep neural
  networks by extrapolation of learning curves.
\newblock In {\em Proc. IJCAI}, 2015.

\bibitem{Efros1999}
A. Efros and T. Leung.
\newblock {Texture Synthesis by non-parametric sampling}.
\newblock In {\em {Proc. Intl. Conf. on Computer Vision (ICCV)}}, 1999.

\bibitem{Elsken2019}
T. Elsken, J. Metzen, and F. Hutter.
\newblock Efficient multi-objective neural architecture search via lamarckian
  evolution.
\newblock In {\em Proc. ICLR}, 2019.

\bibitem{NASSurvey2019}
T. Elsken, J.~H. Metzen, and F. Hutter.
\newblock Neural architecture search: A survey.
\newblock {\em Journal of Machine Learning Research (JMLR)}, 20:1--21, 2019.

\bibitem{Gandelsman2018}
Y. Gandelsman, A. Shocher, and M. Irani.
\newblock Double-dip: Unsupervised image decomposition via coupled
  deep-image-priors.
\newblock In {\em Proc. CVPR}. IEEE, 2018.

\bibitem{Gilbert2018}
A. Gilbert, J. Collomosse, H. Jin, and B. Price.
\newblock Disentangling structure and aesthetics for content-aware image
  completion.
\newblock In {\em Proc. CVPR}, 2018.

\bibitem{GlasnerICCV}
D. Glasner, S. Bagon, and M. Irani.
\newblock Super-resolution from a single image.
\newblock In {\em Intl. Conference on Computer Vision (ICCV)}, 2009.

\bibitem{autogan}
Xinyu Gong, Shiyu Chang, Yifan Jiang, and Zhangyang Wang.
\newblock Autogan: Neural architecture search for generative adversarial
  networks.
\newblock In {\em Proc. ICCV}, 2019.

\bibitem{hays2007scene}
James Hays and Alexei~A Efros.
\newblock Scene completion using millions of photographs.
\newblock In {\em ACM Transactions on Graphics (TOG)}. ACM, 2007.

\bibitem{He2012}
K. He and J. Sun.
\newblock {Statistics of patch offsets for image completion}.
\newblock In {\em {Euro. Conf. on Comp. Vision (ECCV)}}, 2012.

\bibitem{FID}
M. Heusel, H. Ramsauer, T. Unterthiner, B. Nessler, and S. Hochreiter.
\newblock {G}{A}{N}s trained by a two time-scale update rule converge to a
  local nash equilibrium.
\newblock In {\em Proc. NeurIPS}, pages 6629--6640, 2017.

\bibitem{Hutter2019}
Frank Hutter, Lars Kotthoff, and Joaquin Vanschoren, editors.
\newblock {\em Automated Machine Learning: Methods, Systems, Challenges}.
\newblock Springer, 2019.
\newblock In press, available at http://automl.org/book.

\bibitem{IizukaSIGGRAPH2017}
Satoshi Iizuka, Edgar Simo-Serra, and Hiroshi Ishikawa.
\newblock {Globally and Locally Consistent Image Completion}.
\newblock {\em ACM Transactions on Graphics (Proc. of SIGGRAPH 2017)},
  36(4):107:1--107:14, 2017.

\bibitem{Kwatra2003}
V. Kwatra, A. Schodl, I. Essa, G. Turk, and A. Bobick.
\newblock {Graphcut textures: Image and video synthesis using graph cuts}.
\newblock {\em ACM Transactions on Graphics}, 3(22):277--286, 2003.

\bibitem{Liu2018}
C. Liu, B. Zoph, M. Neumann, J. Shelns, W. Hua, L. Li, F-F. Li, A. Yuille, J.
  Huang, and K. Murphy.
\newblock Progressive neural architecture search.
\newblock In {\em Proc. ECCV}, 2018.

\bibitem{Liu2013}
Y. Liu and V. Caselles.
\newblock {Exemplar-based image inpainting using multiscale graph cuts}.
\newblock {\em IEEE Trans. on Image Processing}, pages 1699--1711, 2013.

\bibitem{Miller1989}
G.~F. Miller, P.~M. Todd, and S.~U. Hedge.
\newblock Designing neural networks using genetic algorithms.
\newblock In {\em Proc. Intl. Conf. on Genetic Algorithms}, 1989.

\bibitem{Negrinho2017}
R. Negrinho and G. Gordon.
\newblock Deeparchitect: Automatically designing and training deep
  architectures, 2017.
\newblock arXiv:1704.08792.

\bibitem{pathakCVPR16context}
Deepak Pathak, Philipp Kr\"ahenb\"uhl, Jeff Donahue, Trevor Darrell, and Alexei
  Efros.
\newblock Context encoders: Feature learning by inpainting.
\newblock In {\em CVPR'16}, 2016.

\bibitem{Real2019}
E. Real, A. Aggarwal, Y. Huang, and Q.~V. Le.
\newblock Aging evolution for image classifier architecture search.
\newblock In {\em Proc. AAAI}, 2019.

\bibitem{Real2017}
E. Real, S. Moore, A. Selle, S. Saxena, Y. Suematsu, Q.~V. Le, and A. Kurakin.
\newblock Large-scale evolution of image classifiers.
\newblock In {\em Proc. ICLR}, 2017.

\bibitem{Salimans2016}
T. Salimans, I. Goodfellow, W. Zaremba, V. Cheung, A. Radford, and X. Chen.
\newblock Improved techniques for training {G}{A}{N}s.
\newblock In {\em Proc. NeurIPS}, pages 2234--2242, 2016.

\bibitem{Schulman2017}
J. Schulman, F. Wolski, P. Dhariwal, A. Radford, and O. Klimov.
\newblock Proximal policy optimization algorithms, 2017.
\newblock arXiv:1707.06347.

\bibitem{singan}
Tamar~Rott Shaham, Tali Dekel, and Tomer Michaeli.
\newblock Singan: Learning a generative model from a single natural image.
\newblock In {\em Proc. ICCV}, 2019.

\bibitem{Stanley2002}
K. Stanley and R. Miikkulainen.
\newblock Evolving neural networks through augmenting topologies.
\newblock {\em Evolutionary Computation}, 10:99--127, 2002.

\bibitem{Ulyanov2018}
D. Ulyanov, A. Vedaldi, and V. Lempitsky.
\newblock Deep image prior.
\newblock In {\em The IEEE Conference on Computer Vision and Pattern
  Recognition (CVPR)}, June 2018.

\bibitem{UlyanovJournal}
D. Ulyanov, A. Vedaldi, and V. Lempitsky.
\newblock Deep image prior.
\newblock {\em Intl. Journal Computer Vision (IJCV)}, 2019.

\bibitem{Wang2018}
Ting-Chun Wang, Ming-Yu Liu, Jun-Yan Zhu, Andrew Tao, Jan Kautz, and Bryan
  Catanzaro.
\newblock High-resolution image synthesis and semantic manipulation with
  conditional gans.
\newblock In {\em Proc. CVPR 2018}, 2018.

\bibitem{wang2004imageSSIM}
Zhou Wang, Alan~C Bovik, Hamid~R Sheikh, and Eero~P Simoncelli.
\newblock Image quality assessment: from error visibility to structural
  similarity.
\newblock {\em IEEE transactions on image processing}, 13(4):600--612, 2004.

\bibitem{bam}
M. Wilber, C. Fang, H. Jin, A. Hertzmann, J. Collomosse, and S. Belongie.
\newblock Bam! the behance artistic media dataset for recognition beyond
  photography.
\newblock In {\em Proc. ICCV}, 2017.

\bibitem{wilber2017bamICCV}
Michael~J Wilber, Chen Fang, Hailin Jin, Aaron Hertzmann, John Collomosse, and
  Serge Belongie.
\newblock Bam! the behance artistic media dataset for recognition beyond
  photography.
\newblock {\em arXiv preprint arXiv:1704.08614}, 2017.

\bibitem{DevilDecoder}
Z. Wojna, V. Ferrari, S. Guadarrama, N. Silberman, L. Chen, A. Fathi, and J.
  Uiklings.
\newblock The devil is in the decoder: Classification, regression and gans.
\newblock {\em Intl. Journal of Computer Vision (IJCV)}, 127:1694--1706,
  December 2019.

\bibitem{yeh2016semantic}
Raymond Yeh, Chen Chen, Teck~Yian Lim, Mark Hasegawa-Johnson, and Minh~N Do.
\newblock Semantic image inpainting with perceptual and contextual losses.
\newblock {\em arXiv preprint arXiv:1607.07539}, 2016.

\bibitem{Zhang2019}
H. Zhang, L. Mai, H. KJin, Z. Wang, N. Xu, and J. Collomosse.
\newblock Video inpainting: An internal learning approach.
\newblock In {\em Proc. ICCV}, 2019.

\bibitem{stackgan}
Han Zhang, Tao Xu, Hongsheng Li, Shaoting Zhang, Xiaogang Wang, Sharon Huang,
  and Dimitris Metaxas.
\newblock Stackgan++: Realistic image synthesis with stacked generative
  adversarial networks.
\newblock {\em IEEE Trans. PAMI}, 41(8):1947--1962, 2017.

\bibitem{LPIPS}
R. Zhang, P. Isola, A. Efros, E. Shechtman, and O. Wang.
\newblock The unreasonable effectiveness of deep features as a perceptual
  metric.
\newblock In {\em Proc. CVPR}, 2018.

\bibitem{zhou2016places}
Bolei Zhou, Aditya Khosla, Agata Lapedriza, Antonio Torralba, and Aude Oliva.
\newblock Places: An image database for deep scene understanding.
\newblock {\em arXiv preprint arXiv:1610.02055}, 2016.

\bibitem{Zoph2017}
B. Zoph and Q.~V. Le.
\newblock Neural architecture search with reinforcement learning.
\newblock In {\em Proc. ICLR}, 2017.

\bibitem{Zoph2018}
B. Zoph, V. Vasudevan, J. Shlens, and Q.~V. Le.
\newblock Learning transferrable architectures for scalable image recognition.
\newblock In {\em Proc. CVPR}, 2018.

\end{thebibliography}
}

\end{document}